\documentclass{article}

\usepackage[preprint, nonatbib]{neurips_2023}
\usepackage{graphicx}
\usepackage{amsmath}
\usepackage{amssymb}
\usepackage{booktabs}

\usepackage{url}
\usepackage{multirow}
\usepackage{float}
\usepackage{xspace}
\usepackage{mathtools}
\usepackage{amsfonts}
\usepackage{amsthm}
\usepackage{nicefrac}
\usepackage{enumitem}
\usepackage{caption}

\newcommand{\name}{{\fontfamily{cmss}\selectfont{D4}}\xspace}

\theoremstyle{plain}
\newtheorem{theorem}{Theorem}[section]

\newtheorem{lemma}[theorem]{Lemma}

\theoremstyle{definition}

\theoremstyle{remark}

\usepackage[pagebackref,breaklinks,colorlinks]{hyperref}

\usepackage[capitalize]{cleveref}
\crefname{section}{Sec.}{Secs.}
\Crefname{section}{Section}{Sections}
\Crefname{table}{Table}{Tables}
\crefname{table}{Tab.}{Tabs.}

\usepackage[utf8]{inputenc} 
\usepackage[T1]{fontenc}    
\usepackage{hyperref}       
\usepackage{url}            
\usepackage{booktabs}       
\usepackage{amsfonts}       
\usepackage{nicefrac}       
\usepackage{microtype}      
\usepackage{xcolor}         

\title{D4: \underline{D}etection of Adversarial \underline{D}iffusion \underline{D}eepfakes \\ Using \underline{D}isjoint Ensembles}

\author{%
Ashish Hooda\thanks{~Indicates equal contribution.}$^{1*}$ \quad Neal Mangaokar$^{2*}$ \quad Ryan Feng$^2$ \\ \textbf{Kassem Fawaz}$^1$ \quad \textbf{Somesh Jha}$^1$ \quad
\textbf{Atul Prakash}$^2$\\
$^1$University of Wisconsin-Madison \quad $^2$University of Michigan\\
\texttt{\{ahooda,kfawaz,jha\}@wisc.edu}\\
\texttt{\{nealmgkr,rtfeng,aprakash\}@umich.edu}
}

\begin{document}

\maketitle

\begin{abstract}
Detecting diffusion-generated deepfake images remains an open problem. Current detection methods fail against an adversary who adds imperceptible adversarial perturbations to the deepfake to evade detection. In this work, we propose \textbf{D}isjoint \textbf{D}iffusion \textbf{D}eepfake \textbf{D}etection (\name), a deepfake detector designed to improve black-box adversarial robustness beyond de facto solutions such as adversarial training. \name uses an ensemble of models over disjoint subsets of the frequency spectrum to significantly improve adversarial robustness. Our key insight is to leverage a redundancy in the frequency domain and apply a saliency partitioning technique to disjointly distribute frequency components across multiple models. We formally prove that these disjoint ensembles lead to a reduction in the dimensionality of the input subspace where adversarial deepfakes lie, thereby making adversarial deepfakes harder to find for black-box attacks. We then empirically validate the \name method against several black-box attacks and find that \name significantly outperforms existing state-of-the-art defenses applied to diffusion-generated deepfake detection. We also demonstrate that \name provides robustness against adversarial deepfakes from unseen data distributions as well as unseen generative techniques.

\end{abstract}

\section{Introduction}\label{sec:intro}
Significant advances in deep learning are responsible for the advent of ``deepfakes'', 
which can be misused by bad actors for malicious purposes. Deepfakes broadly refer to digital media that has been synthetically generated or modified by deep neural networks (DNNs). Modern DNNs such as diffusion models~\cite{ddpm,iddpm,ddim,ldm,adm,pndm} and generative adversarial networks (GANs)~\cite{goodfellow2014generative,karras2018progressive,karras2019style,brock2018large,diffusion_projected_gan,projected_gan,choi2020starganv2} are now capable of synthesizing hyper-realistic deepfakes, which can then be used to craft fake social media profiles~\cite{deepfakes_socialmedia}, generate pornography~\cite{vice2017deepfakearticle}, spread political propaganda, and manipulate elections.  

\begin{figure}[t]
     \centering
         \includegraphics[width=\linewidth]{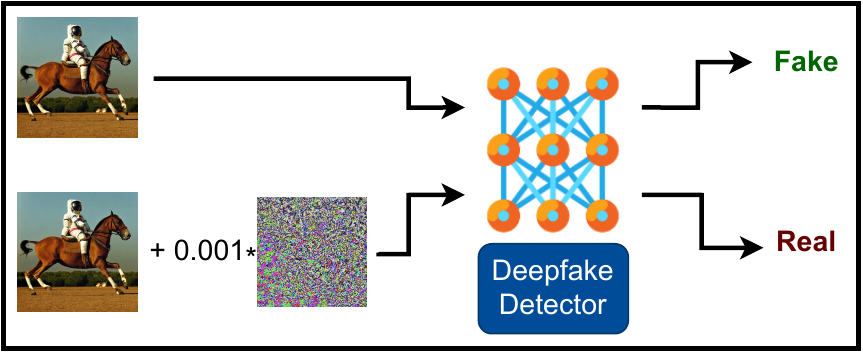}
         \caption{Under the non-adversarial setting, the deepfake detector correctly classifies the image produced by Stable Diffusion~\cite{rombach2022high} on the text prompt “a photograph of an astronaut riding a horse.” as fake. However, one can flip the detector's prediction by adding imperceptible adversarial perturbations to the deepfake.}
         \label{fig:adv_example}
 \end{figure}


The deepfake detection problem asks the defender to classify a given image as deepfake or real. While recent work has made remarkable efforts towards solving the deepfake detection problem, many of these detectors are rendered ineffective by \textit{adversarial examples} (\cref{fig:adv_example}). Specifically, these state-of-the-art detectors often leverage DNNs, and Carlini et al.~\cite{carlini2020evading} (amongst others) have shown that such DNNs are vulnerable --- the attacker can simply use adversarial perturbation techniques to evade detection ~\cite{shahriyar2022evaluating, fernandes2020adversarial, liao2021imperceptible, hussain2022reface}. Recent work has shown that these ``adversarial deepfakes'' can even be crafted in a black-box setting, where the attacker only has query access to the detector~\cite{vo2022adversarial,gandhi2020adversarial, hussain2021adversarial, hussain2022exposing}. Defending against adversarial examples, in general, has been shown to be a difficult task~\cite{athalye2018obfuscated}, and is a critical problem in the deepfake detection setting.


Our key intuition to mitigate this problem is to utilize \textit{redundant information in the frequency feature space} of deepfakes to generate \textit{disjoint ensembles} for adversarial deepfake detection. Specifically, we show in~\cref{sec:redundant_motivation} that we can achieve good detection performance with only a subset of the features, particularly in the frequency domain. This enables us to build an ensemble of performant classifiers, each using a disjoint set of frequencies. In contrast to traditional ensembles (where each model shares the same set of frequencies), a key advantage of this design is that non-robust frequencies are partitioned across all the models in the ensemble (see~\cref{sec:redundancy}). Thus, an attacker is no longer able to perturb a single non-robust frequency to evade all models --- rather, they must find perturbations to evade multiple sets of disjoint frequencies. \name thus aims to thwart the attacker's generation of adversarial deepfakes.


To summarize, our key contributions are as follows:

\begin{enumerate}
\item We propose \name, a diffusion-generated deepfake detection framework designed to be adversarially robust. \name builds an ensemble of models that use disjoint partitions of the input features. This is achieved by leveraging redundancy in the feature space. \name achieves robustness while still maintaining natural deepfake detection average precision scores as high as 93\%. (see~\cref{sec:experimental} for details).

\item Extending the theoretical results by Tramer et al.~\cite{tramer2017space} on dimensionality of adversarial subspaces,  we prove new bounds on the maximum number of adversarial directions that can be found under an ensemble with disjoint inputs. Our bounds are tight for both the $\ell_2$  and $\ell_\infty$ perturbation norms (Lemmas~\ref{lemma:dim_l2} and~\ref{lemma:dim_linf} in~\cref{sec: gaas}) and indicate that \name reduces the dimension of the adversarial subspace, i.e., there are simply fewer adversarial examples to be found.

    \item We evaluate \name against query-based black-box attacks, as well as adaptive frequency and post-processing attacks. Across a variety of diffusion-generated deepfake images, we find that \name significantly outperforms state-of-the-art defenses such as the recently proposed EnsembleDet~\cite{dutta2021ensembledet, wang2020cnn, frank2020leveraging}, suggesting that \name indeed provides a reduction in dimension of the adversarial subspace. For example, as indicated by our evaluation in~\cref{sec:experimental}, \name reduces the attack success rate to 28\%, whereas baselines incur attack success rates of more than 90\%. These improvements also extend to unseen image domains and deepfake generation models.

\end{enumerate}

\section{Background and Related Work}\label{sec:related_work}
\noindent\textbf{Notation.} We consider a distribution $\mathcal{D}$ over $\mathcal{X} \times \mathcal{Y}$, where $\mathcal{X} \subseteq \mathbb{R}^d$ is the input space and $\mathcal{Y} \subseteq \mathbb{Z}^c$ is the finite class-label space. We denote vectors in boldface (e.g., $\mathbf{x}$). We denote a trained classifier as a function $\mathcal{F}: \mathcal{X} \rightarrow\mathcal{Y}$ (the classifier is usually parameterized by its weights $w$, omitted for brevity). We denote the loss function as $\mathcal{L}(\mathbf{x},y)$. An ensemble classifier is a function $M_{(\mathcal{F}_1,\mathcal{F}_2,...,\mathcal{F}_n)}: \mathcal{X} \rightarrow \mathcal{Y}$ that combines the logit outputs $l_1,l_2,...,l_n$ of multiple classifiers $\mathcal{F}_1,\mathcal{F}_2,...,\mathcal{F}_n$ with a voting aggregation function $\mathcal{A}: \mathbb{R}^{n \times c} \rightarrow \mathcal{Y}$. 

For a classifier $\mathcal{F}$ and input-label pair $(\mathbf{x},y)$, an adversarial example is a perturbed input $\mathbf{x'}$ such that (1) $\mathbf{x'}$ is misclassified, i.e., $\mathcal{F}(\mathbf{x'}) \neq y$ and (2) $||\mathbf{x} - \mathbf{x'}||$ is within a small $\epsilon$ ball, where $||.||$ is a given norm. The value of $\epsilon$ is chosen to be small so that the perturbation is imperceptible to humans.

\noindent\textbf{Deepfake Image Generation.} In this work, we focus on detection of deepfake images that are generated entirely from scratch using a deep generative model. Two prominent techniques that have achieved state-of-the-art for such generation are GANs~\cite{goodfellow2014generative} and diffusion models~\cite{ddpm}. GANs comprise two DNNs: a generator and a discriminator. The generator synthesizes images, while the discriminator attempts to distinguish between real and deepfake samples. Through this adversarial training procedure, GANs learn to generate increasingly realistic and high-quality outputs, and have achieved remarkable success~\cite{8667290}. However, GANs have recently been superceded by diffusion models. 
These models iteratively add noise to data samples and then remove it, thereby learning to generate images from randomly sampled noise. Diffusion models have now also achieved new state-of-the-art FID scores for image generation benchmarks~\cite{adm}. Given the quality of images generated, detection of diffusion model deepfakes poses a pressing concern.

\noindent\textbf{Deepfake Image Detection.}
The research community has made rapid progress towards detecting deepfake images. Some efforts propose classification DNNs that operate directly on pixel features~\cite{sha2022fake, wang2020cnn, marra2019incremental, marra2018detection}. Others have  instead trained DNNs using features extracted from the deepfakes, i.e., co-occurrence matrices~\cite{nataraj2019detecting}, color-space anomalies~\cite{mccloskey2018detecting}, convolutional traces~\cite{guarnera2020deepfake}, texture representations~\cite{liu2020global}, pixel-patches~\cite{chai2020makes}, or more recently using neural features extracted from foundation models such as CLIP~\cite{ojha2023towards}.

One particular line of work that has shown great promise for deepfake detection is leveraging \textit{frequency} features. Specifically, Frank et al.~\cite{frank2020leveraging} proposed the idea of detecting deepfakes with the Discrete Cosine Transform (DCT) as a pre-processing transform before the DNN. Similar work has also achieved remarkable performance using frequency features --- ~\cite{zhang2019detecting,durall2019unmasking}, and more recent efforts have emphasized their utility in detecting deepfakes from both GANs and diffusion models~\cite{ricker2022towards}. \name also leverages frequency features, but through a unique disjoint ensembling approach.

\noindent\textbf{Adversarial Deepfakes.} Unfortunately, regardless of the chosen feature space, the aforementioned detectors have been rendered ineffective in adversarial settings. Specifically, Carlini et al.~\cite{carlini2020evading} showed that DNN detectors are vulnerable to adversarial examples --- an adversary can add imperceptible adversarial perturbations to a deepfake that evade such detectors, rendering them ineffective. Others have corroborated this observation~\cite{hussain2021adversarial, hussain2022exposing, neekhara2021adversarial, gandhi2020adversarial, vo2022adversarial, shahriyar2022evaluating, fernandes2020adversarial, liao2021imperceptible, hussain2022reface}.

An adversary can construct these ``adversarial deepfakes'' in either a white-box setting (with complete knowledge of the detector's weights and parameters), or black-box setting (with only query access to the detector). In this work, we focus on the black-box setting. The DNN models of deepfake detectors in the real world are often hidden from the users and likely to be black-box services, such as those already offered by Intel~\cite{intel_detector}, Deepware~\cite{Deepware}, Reality Defender~\cite{RealityDefender} and Sensity AI~\cite{Sensity_2023}. These services are typically available through web-based platforms or through API access. 
Moreover, defending against white-box attacks is a challenging open problem on all vision tasks. We leave defense in the white-box setting to future work.

\noindent\textbf{Adversarial Deepfake Detection.}
Existing defenses for adversarial deepfakes can be broadly classified into (a) training time defenses (which adjust the training process), and (b) inference-time defenses.

\noindent\underline{(a) Training time.} The original training time defense is adversarial training, in which the model is trained on adversarial examples generated during training~\cite{madry2018towards}. However, Carlini et al.~\cite{carlini2020evading} suggest that adversarial training alone is unlikely to achieve significant improvement in robustness in the difficult deepfake detection setting (confirmed by our experiments in ~\cref{sec:ldm_robustness_results}).

Instead of adversarial training, recent work has also proposed using an ensemble of models --- in principle, the adversary is then forced to attack multiple models, instead of just one. However, He et al.~\cite{he2017adversarial} have shown that arbitrarily ensembling models does not necessarily lead to more robustness. Prior work suggests that each model in an ensemble tends to learn the same non-robust features, i.e., an adversary is able to perturb the same features to evade all models~\cite{ilyas2019adversarial}. 

Most recently, Dutta et al.~\cite{dutta2021ensembledet} and Khan et al.~\cite{khan2021adversarially} proposed EnsembleDet and ARDD respectively as ensemble defenses against adversarial deepfakes. Both defenses ensemble multiple different DNN architectures, and posit that the different architectures will learn different features, thereby providing improved robustness. Devasthale et al.~\cite{devasthale2022adversarially} take this approach one step further and also adversarially train each model in the ensemble.

Note that while \name is also an ensemble-based detector, it is fundamentally different than existing approaches since it does not require different architectures to avoid learning the same features --- we instead leverage artifact redundancy to design frequency-partitioned ensembles (see ~\cref{sec:motivation_design} for a more detailed explanation). Furthermore, \name's partitioning approach to achieve disjoint ensembles is novel, and it offers theoretical and empirical advantages where such feature redundancy is available, e.g., generated deepfakes, allowing it to outperform prior work (see ~\cref{sec:experimental}).


\noindent\underline{(b) Inference time.}
Another class of defenses focuses on removing the effect of the adversarial perturbation without any changes made to the underlying detection technique --- unfortunately, these approaches are computationally intensive, e.g., upto 30 minutes per image~\cite{gandhi2020adversarial, jiang2021residual}, in comparison to 
no additional overhead in \name. Nevertheless, both approaches are complementary in that pre-processing could be combined with training-time defenses such as \name.

Finally, we note that there are other inference-time defenses that are not specific to deepfakes. For example, stateful defenses~\cite{li2020blacklight} were proposed to defend against black-box adversarial examples, but have been recently shown to be vulnerable~\cite{feng2023investigating}. Another group of defenses post-process the detector's response by manipulating the detector's confidence scores~\cite{qin2021random, chen2022adversarial}, but these do not work against hard-label black-box attacks.

\section{Our Approach}\label{sec:motivation_design}

\begin{figure*}[t]
     \centering
     \begin{minipage}{.65\textwidth}
         \centering
         \includegraphics[width=\linewidth]{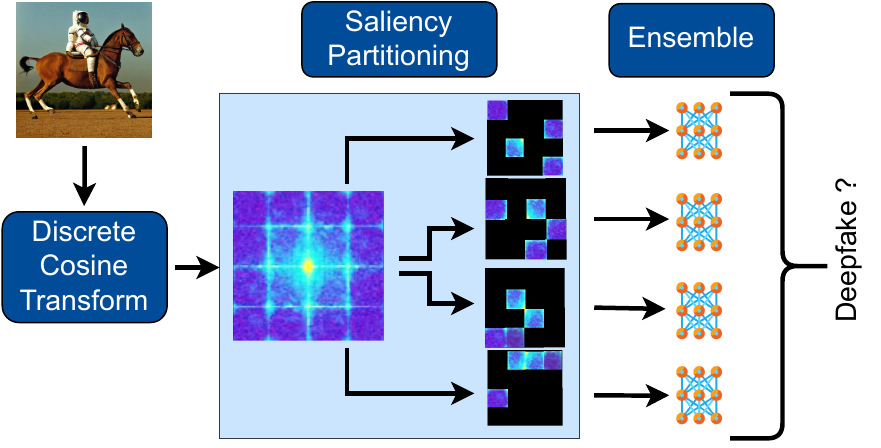}
         \caption{The processing pipeline of \name. It partitions the DCT spectrum of an image into disjoint partitions using a saliency-based approach. Each frequency partition is fed to a separate model that is adversarially trained. A voting mechanism over the ensemble decides the output. }
         \label{fig:pipeline}
     \end{minipage}%
     \hfill
     \begin{minipage}{.3\textwidth}
         \centering
         \includegraphics[width=\linewidth]{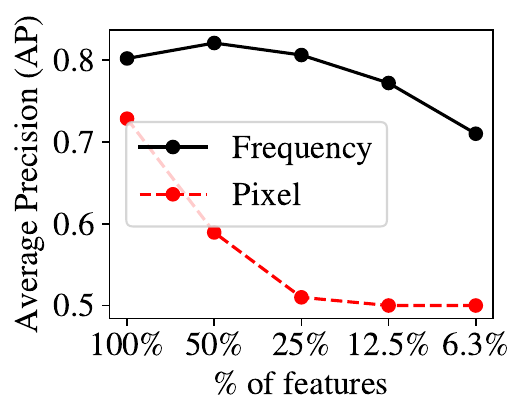}
         \caption{Average Precision (AP) values of a single CNN classifier trained on a fixed subset (randomly selected) of the input features. Redundancy in frequency spectra permits good deepfake detection even when using $\sim$ 6\% of the components (see black line).}
       \label{fig:nat_perf_rand}
     \end{minipage}
     \vspace{-0.1in}
 \end{figure*}

We now present \name (\cref{fig:pipeline}), a deepfake detection framework that leverages an ensemble of disjoint frequency models to achieve robust detection of diffusion-generated deepfake images.~\cref{sec:redundant_motivation} presents our observation of redundant information in the frequency space of deepfakes.~\cref{sec:redundancy} details how redundancy allows frequencies to be partitioned between multiple models for robust ensembling and explains our exact frequency partitioning schemes. Finally,~\cref{sec: gaas} provides theoretical insight into why \name improves adversarial robustness.


\subsection{Redundant Information in Deepfake Image Spectra}\label{sec:redundant_motivation}

As discussed in~\cref{sec:related_work}, ensembles are a promising approach to adversarial robustness, so long as perturbing the same set of features does not simultaneously evade the individual models. We propose designing an ensemble that avoids this shortcoming by \textit{disjointly partitioning the input feature space} amongst individual models. 

As mentioned earlier, the frequency spectrum of images facilitates deepfake detection because generation techniques leave discriminating artifacts that are more prominent in the frequency space. We additionally observe that these artifacts are spread throughout the frequency feature space, and they are more uniformly spread as compared to pixel space. We confirm this in~\cref{fig:nat_perf_rand}, which plots the performance of a simple convolutional classifier~\footnote{We use the same architecture as Frank et al.~\cite{frank2020leveraging}} on an increasingly smaller random subset of the input features.


Our first key insight is thus that disjoint partitioning is feasible for deepfake detection. Specifically, we observe a ``redundancy'' in frequency-space artifacts --- signals relevant for deepfake detection\footnote{Frank et al.~\cite{frank2020leveraging} show that these artifacts manifest in the form of a grid-like spectral pattern, and attribute their presence to the upsampling process in generative models. Existing work proposes detectors that leverage the \textit{entire frequency spectrum} for deepfake detection.} are distributed throughout the frequency spectrum. This observation is best exemplified by the black line in~\cref{fig:nat_perf_rand}, which plots the deepfake detection performance using increasingly small subsets of the spectrum using the same classifier. Using as few as $\sim 6\%$ of the frequency components yields good deepfake detection performance. We emphasize that this does not hold for subsets of pixels (red line), as signals for detection are not well-distributed in the RGB space. Overall, these findings suggest that the frequency-space contains plenty of redundancy, which can potentially be leveraged to design a robust ensemble.

\subsection{Leveraging Redundancy to Build a Robust Ensemble}\label{sec:redundancy}

Using the observations from~\cref{sec:redundant_motivation}, one can craft a robust ensemble by partitioning the frequency components amongst multiple detector models, without hurting natural detection performance.
~\cref{fig:pipeline} visualizes this partitioning as part of our ensembling pipeline. Specifically, for each individual model we mask (i.e., zero out) the frequencies not used. For example, consider an ensemble of two such ``disjoint'' models $F_A$ and $F_B$, with the full-spectrum frequency feature vector $f =[f_1; f_2]$. Then, the input feature vector to $F_A$ is $[f_1; 0]$ and to $F_B$ is $[0; f_2]$. Since the input feature space (frequencies) is not shared amongst the individual models, the adversary cannot simply attack the ensemble by targeting common frequencies.

Furthermore, we note that choice of partitioning scheme, i.e., \textit{how} the frequencies are partitioned plays an important role in robustness of the ensemble. Specifically, the chosen scheme should aim to design all models as ``equals'' --- if some models are less robust than others, then the adversary can target them to overturn the ensemble's decision. 

\noindent\textbf{Saliency Partitioning.} While signals for deepfake detection are distributed throughout the spectrum, there still exists an \textit{adversarial saliency} ordering of these frequencies that determines their robustness for the deepfake detection task. For an ensemble of size $n$, our saliency partitioning technique is aimed at ensuring each model receives a fair proportion of salient frequencies. To this end, we follow~\cite{carlini2017towards} and~\cite{feng2022graphite} to compute saliency values for all frequencies. This is achieved by adversarially perturbing deepfake $x$ to $x + \delta^x$, where $\delta^x$ is the perturbation computed with the Carlini-Wagner $\ell_2$ attack for 1000 steps. Then, we compute saliency $s_i$ for the $i^{th}$ frequency as
\begin{equation}\label{eqn:saliency}
    s_i = \mathbb{E}_{x \in \mathcal{X}} \nabla f(x + \delta^x)_i \cdot \delta^{x}_i
\end{equation}
where subscript $i$ denotes the $i^{th}$ component. Intuitively, higher gradients and larger perturbation magnitudes imply larger saliencies. Frequencies are then ordered by their saliencies, and distributed in a round-robin fashion amongst the models. 


\subsection{Adversarial Subspace of Disjoint Ensembles}\label{sec: gaas}

Given the partitioning approach in~\cref{sec:redundancy}, we now show that an ensemble of such disjoint frequency models increases robustness against adversarial examples by reducing the dimension of the adversarial subspace. For a single model $\mathcal{F}$ and input $\mathbf{x}$, Tramer et al.~\cite{tramer2017space} approximate the $k-$dimensional adversarial subspace as the span of orthogonal perturbations $\mathbf{r_1}, \cdots, \mathbf{r_k} \in \mathbb{R}^d$ such that $\mathbf{r_i}^\intercal \mathbf{g} \geq \gamma  \; \forall \; 1 \leq i \leq k$ where $\mathbf{g} = \nabla_{\mathbf{x}} \mathcal{L}_{\mathcal{F}}(\mathbf{x}, y)$, $\mathcal{L}_{\mathcal{F}}$ is the loss function used to train $\mathcal{F}$, and $\gamma$ is the increase in loss sufficient to cause a mis-classification. For perturbations satisfying the $\ell_2$-norm, i.e. $||\mathbf{r_i}||_2 \leq \epsilon \; \forall \; 1 \leq i \leq k$, the adversarial dimension $k$ is bounded by $\frac{\epsilon^2 ||\mathbf{g}||_2^2}{\gamma^2}$ (tight). In what follows, we extend this result and provide bounds for dimensionality of the shared adversarial subspace between $n$ disjoint models. We provide tight bounds for both $\ell_2$ and $\ell_{\infty}$ norms in Lemma~\ref{lemma:dim_l2} and Lemma~\ref{lemma:dim_linf} respectively (with detailed proofs in Appendix~\ref{ap:proofs}). We consider these bounds for two voting mechanisms: (1) majority, where the ensemble outputs deepfake if at least $\lceil n/2 \rceil$ classifiers predict deepfake, and (2) at-least-one, where the classifier outputs deepfake if at least one classifier predicts deepfake, otherwise it outputs real.

\begin{lemma}\label{lemma:dim_l2}
Given $n$ disjoint models, $\mathcal{F}_1,...,\mathcal{F}_n$, having gradients $\mathbf{g_1}, \cdots, \mathbf{g_n} \in \mathbb{R}^d$ for input-label pair $(\mathbf{x},y)$ (where $\mathbf{g_j} = \nabla_{\mathbf{x}}\mathcal{L}_{\mathcal{F}_j}(\mathbf{x}, y)$), the maximum number $k$ of orthogonal vectors $\mathbf{r_1}, \mathbf{r_2}, \cdots, \mathbf{r_k} \in \mathbb{R}^d$ satisfying $||\mathbf{r_i}||_2 \leq \epsilon$ and,  $\mathbf{r_i}^\intercal \mathbf{g_j} \geq \gamma_j$ for all $1 \leq j \leq n$ (at-least-one voting) or for at least $\lceil\frac{n}{2}\rceil$ models (majority voting), for all $1 \leq i \leq k$ is given by: 
\begin{gather}
    \begin{split}
    k = \min\left(d, \left\lfloor\frac{\epsilon^2 }{\big(\sum\limits_{j=1}^{n}\gamma_j\big)^2} \sum\limits_{j=1}^{n} ||\mathbf{g_j}||^2_2 \right\rfloor\right)  \\ \text{(at-least-one voting)}
    \end{split}\\
    \begin{split}
    k \leq \min\left(d, \left\lfloor
    \max \limits_{|K|=\lceil\frac{n}{2}\rceil} \frac{\epsilon^2}{\big(\sum\limits_{j\in K}\gamma_j\big)^2} \sum\limits_{j\in K} ||\mathbf{g_j}||^2_2 \right\rfloor\right) \\ \text{(majority voting)}
    \end{split}\\
    \begin{split}
    k \geq \min\left(d, \left\lfloor
    \min \limits_{|K|=\lceil\frac{n}{2}\rceil} \frac{\epsilon^2}{\big(\sum\limits_{j\in K}\gamma_j\big)^2} \sum\limits_{j\in K} ||\mathbf{g_j}||^2_2 \right\rfloor\right) \\ \text{(majority voting)}
    \end{split}
\end{gather}
\end{lemma}

\begin{lemma}\label{lemma:dim_linf}
Given $n$ disjoint models, $\mathcal{F}_1,...,\mathcal{F}_n$, having gradients $\mathbf{g_1}, \cdots, \mathbf{g_n} \in \mathbb{R}^d$ for input-label pair $(\mathbf{x},y)$ (where $\mathbf{g_j} = \nabla_{\mathbf{x}}\mathcal{L}_{\mathcal{F}_j}(\mathbf{x}, y)$), the maximum number $k$ of orthogonal vectors $\mathbf{r_1}, \mathbf{r_2}, \cdots, \mathbf{r_k} \in \mathbb{R}^d$ satisfying $||\mathbf{r_i}||_{\infty} \leq \epsilon$ and $\mathbb{E}[\mathbf{g_j}^\intercal \mathbf{r_i}] \geq \gamma_j$ for all $1 \leq j \leq n$ (at-least-one voting) or for at least $\lceil\frac{n}{2}\rceil$ models (majority voting), for all $1 \leq i \leq k$
\begin{equation}
    \begin{split}
    k = \min\left(d, \left\lfloor \min \left( \frac{\epsilon^2||\mathbf{g_1}||_1^2}{n^2 \gamma_1^2},...,\frac{\epsilon^2||\mathbf{g_n}||_1^2}{n^2 \gamma_n^2}\right) \right\rfloor\right) \\ \text{(at-least-one voting)}
    \end{split}
\end{equation}

\begin{equation}
    \begin{split}
    k = \min\left(d, \left\lfloor \text{median} \left( \frac{\epsilon^2||\mathbf{g_1}||_1^2}{n^2 \gamma_1^2},...,\frac{\epsilon^2||\mathbf{g_n}||_1^2}{n^2 \gamma_n^2}\right) \right\rfloor\right) \\ \text{(majority voting)}
    \end{split}
\end{equation}

\end{lemma}

\noindent\textbf{Implications.} If all $n$ disjoint models in the ensemble are ``near-identical" (as expected per our saliency partitioning scheme), i.e., $||\mathbf{g_1}||_2^2 \approx \cdots \approx ||\mathbf{g_n}||_2^2$ and $\gamma_1 \approx \cdots \approx \gamma_n$, then  Lemma 3.1 for at-least-one voting reduces to $k \approx \min\left(d, \left\lfloor\frac{\epsilon^2 ||\mathbf{g_1}||^2_2 }{n\gamma_1^2}  \right\rfloor\right)$. This implies that an ensemble of $n$ disjoint models offers potential reduction in dimensionality of the adversarial subspace by a factor of $n$ compared to any individual constituent disjoint model. Similar interpretation holds for Lemma 3.2, where reductions are now by a factor of $n^2$. Next, in~\cref{sec:experimental}, we empirically demonstrate that this reduction in dimension of adversarial subspace leads to improved performance against black-box adversarial examples.

\section{Experimental Evaluation}\label{sec:experimental}
Our experiments broadly aim to answer the following questions:

{\noindent\textbf{Q1.}} How robust are \name ensembles against black-box attackers, and how does \name raise the attacker's cost of creating adversarial deepfakes? 

\noindent\textbf{Q2.} Does \name's robustness hold for deepfakes from different diffusion models? How well does \name generalize to deepfakes
from unseen image domains, diffusion models, or even different generative architectures, i.e., GANs?

\noindent\textbf{Q3.} How does \name fare against traditional post-processing, and more adaptive frequency-based attacks?

In the following sections, we address these questions by detailing our setup and experiments.

\subsection{Experimental Setup}\label{sec: exp_setup}
We adopt the following settings to evaluate \name for detecting deepfakes and adversarial deepfakes.

\begin{figure*}[]
\scriptsize
\centering
\begin{tabular}{@{}cccccccc@{}}
\toprule
\multicolumn{1}{c}{\multirow{2}{*}{\textbf{Detector}}} & \multicolumn{1}{c}{\multirow{2}{*}[0.1em]{\textbf{No Attack}}} & \multicolumn{6}{c}{\textbf{Attack (ASR)}}\\
\cmidrule{3-8}
& \multirow{1}{*}[0.05em]{\textbf{(AP)}} & \textbf{SurFree} & \textbf{HSJA} & \textbf{QEBA} & \textbf{Triangle} & \textbf{Boundary} & \textbf{SignOPT} \\  \midrule
\textbf{EnsembleDet} &     100\%                                                              &      100\%            &   100\%             &    100\%            &    77\%               &        100\%            &       100\%          \\ 
\textbf{ARDD} &       100\%                                                            &    {100\%}              &      {100\%}         &       {100\%}        &        {99\%}           &      90\%             &     100\%           \\
\textbf{ARDD-AT} &     100\%                                                              &      91\%            &     23\%          &     97\%          &      18\%             &     19\%              &     43\%           \\
\textbf{\name  (SIZE=1)} &       98\%                                                                    &        {93\%}          &       11\%        &    53\%           &    {51\%}               &     \textbf{6\%}              &    10\%            \\
\textbf{\name  (SIZE=4)} &   93\%    &    \textbf{28\%}                                                              &       \textbf{3\%}           &    \textbf{2\%}           &         \textbf{0\%}      &    8\%               &    \textbf{8\%}                             \\ \bottomrule
\end{tabular}
\captionof{table}{An attacker achieves lower attack success rates (ASRs) when attacking \name (SIZE=4) as compared to the baselines. Attacks are launched on LSUN bedroom deepfake images from an LDM diffusion model, with a query budget of 50k queries and $\ell_2$ perturbation budget of $\epsilon=10$.}
\label{tab:ldm_robust_recall_all_detectors}

\end{figure*}

\begin{figure*}[]
\scriptsize
\centering
    \includegraphics[width=0.5\columnwidth]{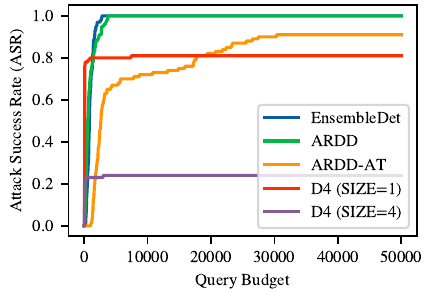}
    \captionof{figure}{SurFree ASR vs. query budget against all detectors, on LSUN bedroom deepfakes from the LDM diffusion model.}
    \label{fig:budget_vs_asr}

\end{figure*}

\noindent\textbf{Datasets and Pre-processing.} We perform our experiments using real images from the LSUN Bedroom dataset~\cite{lsun}, and the CelebaHQ dataset~\cite{karras2018progressive}. For each of these datasets, we obtain deepfake images from the LDM~\cite{ldm}, DDIM~\cite{ddim}, and PNDM~\cite{pndm} diffusion models. Deepfakes for bedroom are sourced from those already generated by prior work~\cite{ricker2022towards}, and we generate deepfakes for CelebaHQ ourselves using models from the HuggingFace public model repository~\cite{wolf2019huggingface}. For our generalization evaluation with GANs, we source ProGAN~\cite{karras2018progressive}, StyleGAN~\cite{karras2019style}, and Diff-StyleGAN2~\cite{diffusion_projected_gan} images again from~\cite{ricker2022towards}. For any given diffusion model and dataset, the training set comprises 39k training images for each of the deepfake and real classes, 1k validation, and 10k testing (as per~\cite{ricker2022towards}). All images are resized to 256x256, and then center-cropped to 224x224. 

\noindent\textbf{Baselines.} 
We implement five baselines to evaluate \name.
The first two baselines are the pixel-space ensemble defenses \textit{EnsembleDet}~\cite{dutta2021ensembledet} and \textit{ARDD}~\cite{khan2021adversarially} proposed by prior work. EnsembleDet comprises three models with the EfficientNet, XCeption, and Resnet50 architectures. ARDD is similar, but instead comprises the VGG16, InceptionV3, and Xception architectures. Our third baseline \textit{ARDD-AT} is also from prior work~\cite{devasthale2022adversarially}, which changes the architectures of ARDD to VGG19, Vision Transformer, Wide-ResNet, and also adversarially trains each individual model using the PGD-150 attack on deepfake images only. Our fourth baseline is simply an adversarially trained version of a full-spectrum frequency space detector --- this is equivalent to a \name ensemble of size 1, i.e., \textit{\name (SIZE=1)} without multiple models or disjoint partitioning. We perform adversarial training for \name (SIZE=1) with PGD-50 attacks~\cite{madry2018towards} using the TRADES loss~\cite{zhang2019theoretically}. Finally, we use the pretrained CNNDet detector from~\cite{wang2020cnn} as an additional baseline for our generalization experiments in~\cref{sec:unseen_generative}, since this detector was designed for detection of unseen generative models~\footnote{The pre-trained models for the more recent generalization detector from~\cite{ojha2023towards} are currently unavailable.}. All baseline models (except CNNDet) are trained using the Adam optimizer~\cite{kingma2014adam} with an initial learning rate of 0.0001, batch size of 32, and a maximum of 20 epochs.

\noindent\textbf{Attacks.} We evaluate \name and baselines against six popular black-box attacks spanning both the gradient-based and random search-based categories. For gradient-based attacks, we select HSJA~\cite{hsja} and QEBA~\cite{qeba}. For random search-based attacks, we select SurFree~\cite{surfree}, Triangle~\cite{triangleattack}, Boundary~\cite{boundary}, and SignOPT~\cite{signopt}. We employ the default hyperparameters for each attack, and focus our attack evaluation on the $\ell_2$ norm with a standard perturbation $\epsilon=10$. We impose a query budget of 50,000 queries on each attack following~\cite{qin2021random} (roughly equivalent to between \$50 - \$75 as per modern MLaaS platforms such as Clarifai).

\noindent\textbf{\name Architecture and Training.} 
We implement a \name ensemble \textit{\name (SIZE=4)} comprising four models. Each model follows a ResNet50 architecture, and the 2D-Discrete Cosine Transform (DCT) is used to convert images to the frequency space before distributing frequencies amongst the models. Each individual model is also adversarially trained using the same procedure as the \name (SIZE=1) baseline.


\begin{table*}[]
\scriptsize
\centering
    \begin{tabular}{@{}cccc@{}}
\toprule
\multirow{2}{*}{\textbf{Model}}                 & \multirow{2}{*}{\textbf{Detector}} & \multicolumn{1}{c}{\multirow{2}{*}{\textbf{\begin{tabular}[c]{@{}c@{}}No Attack\\ (AP)\end{tabular}}}} & \multirow{2}{*}{\textbf{\begin{tabular}[c]{@{}l@{}}Attack\\ (ASR)\end{tabular}}} \\
                                                  &                                  & \multicolumn{1}{c}{}                                                                                   &                                                                                     \\ \midrule
\multicolumn{1}{c}{\multirow{2}{*}{\textbf{LDM}}} & \textbf{\name (SIZE=1)}                      & 98\%                                                                                              &    95\%                                                                                 \\ 
\multicolumn{1}{c}{}                              & \textbf{\name (SIZE=4)}                 & 93\%                                                                                              &           \textbf{28\%}                                                                          \\ \midrule
\multirow{2}{*}{\textbf{DDIM}}                    & \textbf{\name (SIZE=1)}                      & 98\%                                                                                              &       97\%                                                                              \\ 
                                                  & \textbf{\name (SIZE=4)}                 & 78\%                                                                                              &    \textbf{15\%}                                                                                 \\ \midrule
\multirow{2}{*}{\textbf{PNDM}}                    & \textbf{\name (SIZE=1)}                      &                      97\%                                                                                  &      100\%                                                                               \\  
                                                  & \textbf{\name (SIZE=4)}                 &       97\%                                                                                                 &            \textbf{0\%}                                                                         \\ \bottomrule
\end{tabular}
\caption{\name (SIZE=4) only trained on LSUN Bedroom is able to generalize its robustness to CelebaHQ, an entirely different domain unseen during training time.}
\label{tab: generalize_celebahq}

\end{table*}

\begin{table*}[]
\scriptsize
\centering
    \begin{tabular}{@{}cccccccc@{}}
\toprule
\multirow{2}{*}[-0.1em]{\textbf{Model}}                 & \multirow{2}{*}[-0.1em]{\textbf{Detector}} & \multicolumn{6}{c}{\textbf{Attack (ASR)}}                                                                       \\ \cmidrule(l){3-8} 
                                                  &                                  & \textbf{SurFree} & \textbf{HSJA} & \textbf{QEBA} & \textbf{Triangle} & \textbf{Boundary} & \textbf{SignOPT} \\ \midrule
\multicolumn{1}{c}{\multirow{2}{*}{\textbf{LDM}}} & \textbf{\name (SIZE=1)}                      & {93\%}          &       11\%        &    53\%           &    {51\%}               &     \textbf{6\%}              &    10\%            \\  
\multicolumn{1}{c}{}                              & \textbf{\name (SIZE=4)}                 & \textbf{28\%}                                                              &       \textbf{3\%}           &    \textbf{2\%}           &         \textbf{0\%}      &    8\%               &    \textbf{8\%}                             \\\midrule
\multirow{2}{*}{\textbf{DDIM}}                    & \textbf{\name (SIZE=1)}                      & 95\%        & 18\%     &  99\%    &  \textbf{4\%}        & 26\%         & 24\%      \\  
                                                  & \textbf{\name (SIZE=4)}                 & \textbf{9\%}        & \textbf{13\%}     & \textbf{5\%}     & 6\%         & \textbf{11\%}         & \textbf{15\%}      \\ \midrule
\multirow{2}{*}{\textbf{PNDM}}                    & \textbf{\name (SIZE=1)}                      & 100\%        & 87\%     & 77\%     & 16\%         & 91\%         & 89\%      \\  
                                                  & \textbf{\name (SIZE=4)}                 & \textbf{0\%}        & \textbf{0\%}     & \textbf{0\%}     & \textbf{0\%}         & \textbf{0\%}         & \textbf{0\%}      \\ 
\bottomrule
\end{tabular}
\caption{\name (SIZE=4) continues to reduce the attacker's ASR more than the baselines, even when trained and tested on adversarial deepfakes from diffusion models other than LDM.}
\label{tab:ddim_pndm_robustness}

\end{table*}

\noindent\textbf{Metrics.} We follow prior work~\cite{wang2020cnn,ricker2022towards,ojha2023towards} and use average precision (AP) to measure the natural, i.e., non-adversarial, unperturbed deepfake detection performance of \name. We then consider an adversary that attempts to perturb deepfakes to the ``real'' class, and employ \textit{attack success rate}, i.e., ASR (fraction of successfully perturbed deepfakes) as our performance metric for robustness. Lower ASR implies that the detector is more effective against adversarial deepfakes.
\begin{table*}[]
\scriptsize
\centering
\begin{tabular}{@{}ccccccc@{}}
\toprule
\textbf{Detector}                      & \textbf{LDM} & \textbf{DDIM} & \textbf{PNDM} & \textbf{StyleGAN} & \textbf{ProGAN} & \textbf{Diff-StyleGAN2} \\ \midrule
\textbf{CNNDet}                        & 62\% (-)     & 68\% (100\%)  & 64\% (100\%)  & \textbf{95} (100\%)        & \textbf{100\%} (100\%)   & \textbf{100\%} (100\%)           \\
\textbf{\name (SIZE=4)} & \textbf{93\% (28\%)}   & \textbf{70\%} (\textbf{9\%})    & \textbf{79\% (19\%)}   & 67\% (\textbf{14\%})       & 59\% (28\%)      & 67\% (11\%)             \\
\textbf{Both}                                   & \textbf{93\% (28\%)}  & \textbf{70\% (9\%)}    & \textbf{79\% (19\%)}   & 80\% (54\%)       & \textbf{100\%} (\textbf{63\%})    & 94\% (\textbf{67\%})             \\ \bottomrule
\end{tabular}
\caption{Generalization of CNNDet (trained on ProGAN) and \name (SIZE=4) (trained on LDM) to diffusion and GAN deepfakes that were unseen during training. Results are presented in the following format: non-adversarial AP (ASR). }
\label{tab:unseen_model_generalization}
\end{table*}

\subsection{Robustness Against Adversarial Examples}\label{sec:ldm_robustness_results}
We now present performance of \name and baselines under the six attacks described in ~\cref{sec: exp_setup}. For each detector, we present ASR over 100 images under a 50k query budget.

Results for each baseline detector are presented in rows 1-4 of~\cref{tab:ldm_robust_recall_all_detectors}. Notably, these baselines achieve excellent AP scores of $\sim 100\%$ on non-adversarial deepfakes. However, we find that for each detector at least one attack achieves ASR $> 90$\%, rendering it entirely ineffective. Interestingly, the random-search based SurFree attack is particularly effective against all the baselines, e.g., 91\% and 93\% against the the ARDD-AT and \name (SIZE=1) baselines respectively. In contrast \name (SIZE=4) (presented in row 5) is not vulnerable to any ASRs over 28\% (again, achieved by SurFree). In some cases, it can even reduce ASR to $< 3$\%. Furthermore, \name is able to withstand these attacks without much drop in performance on non-adversarial deepfakes (93\% AP). We re-emphasize that this is only achievable due to the redundancy observation from ~\cref{sec:redundant_motivation}.

To better visualize how \name raises the cost of an attack, we also plot SurFree ASR against attack query budget for all detectors in~\cref{fig:budget_vs_asr}. An attacker can typically achieve over 80\% ASR against all baselines within 20k queries. However, \name (SIZE=4) continues to prevent ASR over 28\% even at over double, i.e. 50k queries.

\subsection{Generalization of Robustness}
We now expand beyond the standard setting discussed in~\cref{sec:ldm_robustness_results} and instead consider the generalization capabilities of \name in different contexts. First, we repeat the experiments from~\cref{sec:ldm_robustness_results} for other diffusion models. Second, we consider a more difficult setting and extend our evaluation of \name to adversarial deepfakes from models and domains \textit{unseen at training time}. While this generalization is known to be possible for non-adversarial deepfake detection~\cite{wang2020cnn,ojha2023towards}, to the best of our knowledge generalization for adversarial deepfake detection has not yet been explored. 

\subsubsection{Different Diffusion Models}
~\cref{tab:ddim_pndm_robustness} presents the results of repeating the experiments in ~\cref{sec:ldm_robustness_results}, but now for adversarial deepfakes from the DDIM and PNDM diffusion models. As expected, we observe that the trends continue to hold --- in fact, \name (SIZE=4) on PNDM is able to completely prevent all six attacks, i.e., ASR=0 across all images. For reference, the \name (SIZE=1) baseline (one of the stronger ones from~\cref{tab:ldm_robust_recall_all_detectors}) is again  vulnerable to at least one attack with $> 90\%$ ASR.

\subsubsection{Unseen Image Domains}
~\cref{tab: generalize_celebahq} presents the results of evaluating the \name (SIZE=4) models from~\cref{tab:ddim_pndm_robustness} (which were trained only on bedroom images) on adversarial deepfakes from an entirely different data domain, i.e., human faces. We focus on the strongest SurFree attack. Again, \name reduces ASR significantly more than the baselines, with minimal cost to non-adversarial deepfake detection. One exception is DDIM, for which non-adversarial AP drops to 78\%. However, the decrease in an attacker's ASR from 97\% to 15\% likely offsets this drop in adversarial settings.

On potential explanation for \name's success over the baselines here is as follows: any detector using the full feature set may overfit to a small set of non-robust features unique to the specific training image domain. On the other hand, \name's disjoint partitioning of non-robust features forces each model to learn more from the robust artifacts caused by the diffusion model. Overall, \name suggests that generalizing adversarial robustness to a variety of domains is possible.

\subsubsection{Unseen Generative Models}\label{sec:unseen_generative}
We now evaluate \name's generalization to adversarial deepfakes from models unseen during training time, including different diffusion models, as well as other architectures, i.e., GANs. To this end, we select the \name (SIZE=4) model from~\cref{tab:ldm_robust_recall_all_detectors} trained on LSUN bedroom only. Since prior work has only focused on generalization in the non-adversarial context, our baseline is the popular CNNDet detector~\cite{wang2020cnn} renown for its detection capabilities across a wide variety of GANs. CNNDet is a pixel-space detector trained only on ProGAN deepfake images, using heavy JPEG and blurring data augmentation. 

Row 1 of~\cref{tab:unseen_model_generalization} presents CNNDet's AP scores for non-adversarial deepfakes and ASR for adversarial deepfakes, across the variety of diffusion and GAN models listed in~\cref{sec: exp_setup}. As expected, CNNDet performs well for detection of non-adversarial GAN deepfake images (all AP scores $> 80\%$). However, it performs poorly for diffusion deepfakes which can be explained by prior work's observation that the high frequency artifacts differ between GANs and diffusion models~\cite{ricker2022towards}. Under the adversarial setting, it is rendered completely ineffective with 100\% ASR for both GAN and diffusion deepfakes. This also suggests that simultaneously acheiving both generalization and robustness against adversarial examples is a challenging problem. Row 2 presents \name (SIZE=4) scores, which exhibit the opposite trend --- it is able to generalize better (both non-adversarial and adversarial) for diffusion deepfakes, but worse for GANs (since it is trained on diffusion images).

The above observations suggest that \textit{combining the two detectors} may yield improvements. To this end, row 3 presents the results of ensembling \name (SIZE=4) and CNNDet using an at-least-one voting scheme. The resulting aggregate detector ``merges'' the benefits to an extent, presenting AP scores $>=80\%$ and $>=70\%$ for non-adversarial GAN and diffusion model deepfakes respectively. Furthermore, ASRs are significantly reduced from the 100\% of CNNDet. Overall, this indicates that \name is complementary to existing detectors, and can be combined to improve adversarial (or even non-adversarial) generalization.

\subsection{Post-Processing and Adaptive Attacks}
We now evaluate \name (SIZE=4)'s robustness   to standard post-processing image transforms that are common, e.g., when distributed through social media. We focus on standard transforms leveraged by prior work, including additive Gaussian noise ($\sigma=2$), blurring ($\sigma=2$), and JPEG compression (80\%). We also evaluate against the adaptive frequency-peaks attack proposed by recent work~\cite{wesselkamp2022misleading}, designed to evade frequency-based deepfake detectors by removing frequency artifacts. At a high level, the attack manipulates frequency coefficients to remove ``peaks" in the spectrum. Specifically, it computes a fingerprint as the difference between the log-scaled mean spectra of deepfake and real images. To attack a deepfake image, the attack then intensifies and subtracts this fingerprint from the image. 

~\cref{tab:post_processing} presents results of evaluating \name (SIZE=4) under the above settings. We observe that \name is generally robust to post-processing, with the largest drop happening for Gaussian noise (78\% AP). This is to be expected, as prior work has shown that the frequency space is generally robust to all standard transforms except noise~\cite{frank2020leveraging}. Furthermore, the frequency-peaks attack does not appear to hurt performance. This is likely because a disjoint partitioning of features ensures that many frequency components are used for detection, and not just the peaks.

\begin{table}[]
\scriptsize
\centering
\begin{tabular}{ccccc}
\toprule
\textbf{Detector} & \textbf{Noise} & \textbf{Blur} & \textbf{JPEG} & \textbf{Freq-Peaks} \\ \midrule
\textbf{D4 (SIZE=4)}  & 78\% & 93\% & 83\% & 92\% \\\bottomrule
\end{tabular}
\caption{AP scores on LSUN Bedroom LDM deepfakes that are subject to post-processing, or an adaptive frequency-peaks attack.}
\label{tab:post_processing}
\end{table}

\section{Discussion and Future Work}
\label{sec:discussion}

\begin{figure}[t]
        \centering
        \includegraphics[width=0.5\linewidth]{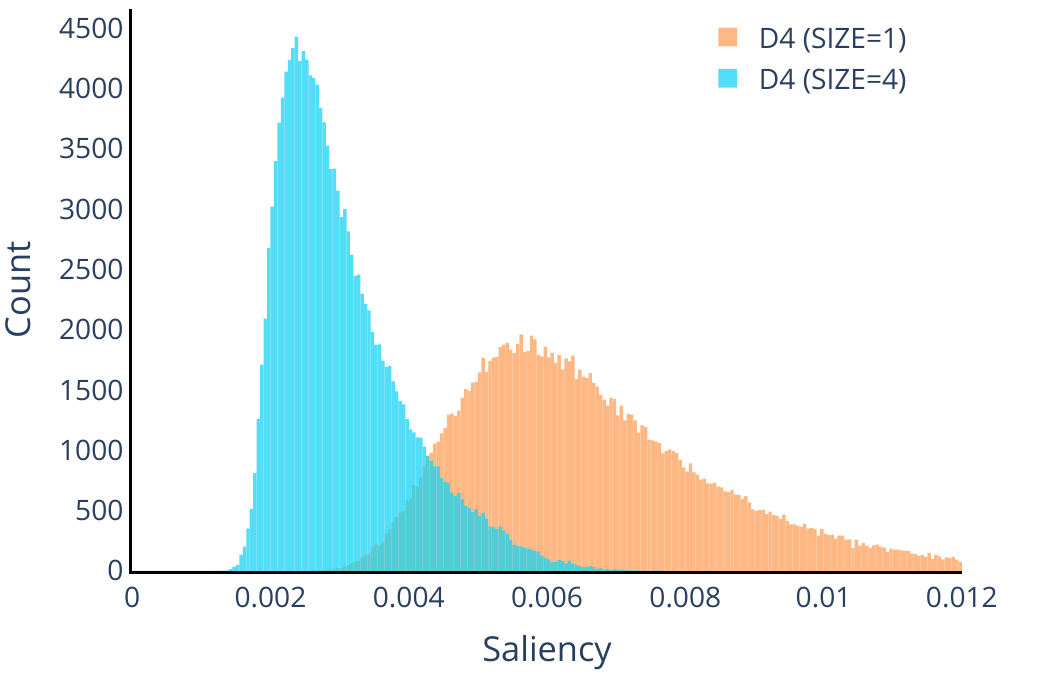}
    \caption{Distribution of frequency feature saliencies for \name (SIZE=1) and \name (SIZE=4) ensembles. Lower saliency implies higher feature robustness. }
    \label{fig:freq_sal_dist}
\end{figure}

\noindent\textbf{Societal impact and limitations.} Modern deepfakes raise several societal and security threats; \name is a step towards mitigating that. Nonetheless, adversarial deepfakes also have benign use cases, e.g., anonymization of an end-user on an online network; \name could prevent such anonymization. Additionally, \name is focused on diffusion-generated deepfake images --- since it relies upon redundancy in the frequency space, it may not be effective against against future types of deepfakes that avoid these artifacts. We believe that the benefits of \name outweigh such potential concerns. 

\noindent\textbf{Alternate Partitioning Strategies.}
Recall from Section~\ref{sec:redundancy} that saliency of a feature may be viewed as a heuristic measure of its ``robustness", as larger saliencies imply that the model is more sensitive to perturbations of that frequency. Figure~\ref{fig:freq_sal_dist} plots the distribution of absolute saliency values for \name (SIZE=1) and \name (SIZE=4) ensembles. We observe that saliencies for saliency-partitioning configurations \name (SIZE=4) are of relatively lower values, and are \textit{sharply concentrated around their mode}, implying higher feature robustness. This can be attributed to the round-robin, equal distribution of robust frequencies amongst the constituent models. Improved approaches to saliency partitioning could increase this separation, improving model robustness even further. We leave this exploration to future work.

\noindent\textbf{Applicability to domains other than deepfake detection.} We presented \name as a framework for adversarially robust deepfake detection. However, we hypothesize that this approach could apply to other classification tasks that exhibit redundancy in a feature space. While we are unaware of such a space for the popular CIFAR10 and ImageNet classification tasks
, there are several classification tasks in, say, the audio domain that exhibit redundancy in features, e.g., keyword spotting and fake speech detection. Exploring this hypothesis is an interesting future research direction.

\section{Conclusions}
In this paper, we present \name, an ensemble approach to deepfake detection that exploits redundancy in frequency feature space by partitioning the frequencies across multiple models. We show theoretical advantages to such disjoint partitioning of input features, that reduces the dimensionality of the adversarial subspace. We empirically validate that \name offers significant gains in robustness under black-box attacks, reducing attack success rates to as low as 0\%.



{\small
\bibliographystyle{unsrt}
\bibliography{main}
}

\clearpage
\section{Appendix}
In Appendix~\ref{ap:proofs}, we provide proofs for our bounds on dimensionality of the adversarial subspace of disjoint ensembles, such as \name. 
In Appendix~\ref{ap:eval}, we provide details of our baseline configurations and attack settings for reproducibility purposes, as well as additional experimental results on other datasets. 

\subsection{Proofs}\label{ap:proofs}
\subsubsection{Orthogonal Gradients}\label{ap:proof_orthogonal}
Prior to providing proofs for the adversarial dimensions, we demonstrate that gradients for disjoint classifiers are always orthogonal to each other. We will use this for our later results. Given input space $\mathcal{X}$ and class-label space $\mathcal{Y}$, we have $n$ disjoint classifiers $\mathcal{F}_1,...,\mathcal{F}_n$. If $T$ is the DCT transformation matrix, we can define $T_i$ to be the transformation matrix for the classifier $\mathcal{F}_i$. Each disjoint transformation $T_i$ has a lot of zeros. Only the rows corresponding to the unmasked frequencies of classifier $\mathcal{F}_i$ have non-zero entries. Moreover, since no frequency is shared by any two classifiers, the $j^{th}$ row will have non-zero entries in exactly one of the $n$ disjoint transformation matrices, i.e. $T_i T_j^\intercal = O \; \forall i \neq j$.

Next, the $n$ disjoint classifiers $\mathcal{F}_1,...,\mathcal{F}_n$, where $\mathcal{F}_i : T_i\mathbf{x} \rightarrow y$, are trained using loss functions $\mathcal{L}_{\mathcal{F}_1},...,\mathcal{L}_{\mathcal{F}_n}$ respectively. Now, the dot product between the gradients of classifiers $\mathcal{F}_i$ and $\mathcal{F}_j$ is given by

\begin{equation}
\label{eq:ortho}
\begin{split}
    \left(\nabla_{\mathbf{x}} \mathcal{L}_{\mathcal{F}_i}\right)^\intercal \left(\nabla_{\mathbf{x}} \mathcal{L}_{\mathcal{F}_j}\right) &= \left(T_i^\intercal \nabla_{T_i \mathbf{x}}\mathcal{L}_{\mathcal{F}_i}\right)^\intercal \left(T_j^\intercal \nabla_{T_j \mathbf{x}}\mathcal{L}_{\mathcal{F}_j}\right) \\
    &= \left(\nabla_{T_i \mathbf{x}}\mathcal{L}_{\mathcal{F}_i}\right)^\intercal T_i T_j^\intercal \left(\nabla_{T_j \mathbf{x}}\mathcal{L}_{\mathcal{F}_j}\right) \\
    &= 0
\end{split}
\end{equation}

\subsubsection{Proof of Lemma 3.1}\label{ap:proof_l2}
From~\cite{tramer2017space}, we know that for a classifier $\mathcal{F}:\mathcal{X} \rightarrow \mathcal{Y}$ where $\mathcal{X} \in \mathbb{R}^d$ is the input space and $\mathcal{Y}^c \in \mathbb{Z}$ is the finite class label space, the dimension of the adversarial subspace around input-label pair $(\mathbf{x},y)$ where $\mathbf{x} \in \mathcal{X}$ and $y \in \mathcal{Y}$, is approximated by the maximal number of orthogonal perturbations $\mathbf{r_1}, \mathbf{r_2}, ..., \mathbf{r_k}$ such that $||\mathbf{r_i}||_2 \leq \epsilon$ and $\mathbf{g}^\intercal \mathbf{r_i} \geq \gamma \; \forall \; 1\leq i \leq k$. Here, $\mathbf{g} = \nabla_{\mathbf{x}} L(\mathcal{F}(\mathbf{x}),y)$ and $\gamma$ is the increase in loss function $L$ sufficient to cause a mis-classification. ~\cite{tramer2017space} provide a tight bound for $k$:
\begin{equation}
\label{eq:tramer}
    k = \min\left(d, \left\lfloor\frac{\epsilon^2 ||\mathbf{g}||^2_2}{\gamma^2} \right\rfloor\right)
\end{equation}

We now extend this result for $n$ disjoint classifiers. Let $\mathbf{g'} = \frac{\sum\limits_{j=1}^{n} \mathbf{g_j}}{n}$. \\
Now, for \textit{at-least-one voting}, 

\begin{equation}
\label{eq:ineql2one}
\mathbf{g'}^\intercal \mathbf{r_i} = \frac{\sum\limits_{j=1}^{n} \mathbf{g_j}^\intercal \mathbf{r_i}}{n} \geq \frac{\sum\limits_{j=1}^{n} \gamma_j}{n} \quad \forall \; 1 \leq i \leq k
\end{equation}

Applying the result from~\cite{tramer2017space} (Equation~\ref{eq:tramer}) on the above inequality (Equation~\ref{eq:ineql2one}), we get:

\begin{equation}
\label{eq:l2}
\begin{split}
k & = \min\left(d, \left\lfloor\frac{\epsilon^2 n^2 ||\mathbf{g'}||^2_2 }{\left(\sum\limits_{j=1}^{n} \gamma_j\right)^2} \right\rfloor\right) \\
& = \min\left(d, \left\lfloor\frac{\epsilon^2 \sum\limits_{j=1}^{n}||\mathbf{g_j}||^2_2}{\left(\sum\limits_{j=1}^{n} \gamma_j\right)^2} \right\rfloor\right).  \\ & \text{(since $\mathbf{g_i}^\intercal \mathbf{g_j} = 0 \;~\forall i \neq j$, using Equation~\ref{eq:ortho})} 
\end{split}
\end{equation}

Now, for \textit{majority voting}, we again apply the results from~\cite{tramer2017space} (Equation~\ref{eq:tramer}). However, the derivation now depends on the selection of $\left\lceil\frac{n}{2}\right\rceil$ models that the adversary chooses to target. To obtain the lower and upper bounds, we can select $\left\lceil\frac{n}{2}\right\rceil$ with the most and least adversarial dimensions respectively. Following a similar derivation as before, we get :

\begin{equation}
\label{eq:l2}
\begin{split}
 k & \geq \min\left(d, \left\lfloor \min \limits_{|K|=\lceil\frac{n}{2}\rceil} \frac{\epsilon^2 \sum\limits_{j=1}^{n}||\mathbf{g_j}||^2_2}{\left(\sum\limits_{j=1}^{n} \gamma_j\right)^2} \right\rfloor\right)
\end{split}
\end{equation}

\begin{equation}
\label{eq:l2}
\begin{split}
 k & \leq \min\left(d, \left\lfloor \max \limits_{|K|=\lceil\frac{n}{2}\rceil} \frac{\epsilon^2 \sum\limits_{j=1}^{n}||\mathbf{g_j}||^2_2}{\left(\sum\limits_{j=1}^{n} \gamma_j\right)^2} \right\rfloor\right)
\end{split}
\end{equation}

\subsubsection{Proof of Lemma 3.2}\label{ap:proof_linf}
Follow up work from~\cite{tramer2017ensemble} also provides a tight bound for the adversarial dimension in the $\ell_{\infty}$ case. They provide a tight bound for the number of $k$ orthogonal perturbations $\mathbf{r_1},...,\mathbf{r_k} \in \mathbb{R}^d$ such that $||\mathbf{r_i}||_{\infty} \leq  \epsilon$, given by $sign(\mathbf{g})^\intercal \mathbf{r_i} = \frac{\epsilon d}{\sqrt{k}} \; \forall 1 \leq i \leq k$ where $sign(\mathbf{g})$ is the signed gradient. 

We now extend this result for $n$ disjoint classifiers. For $\mathbf{g'} = \frac{\sum\limits_{j=1}^{n} \mathbf{g_j}}{n}$, since $\mathbf{g_j}'s$ are non-zero only on non-overlapping dimensions, we can see that $sign(\mathbf{g'})^\intercal r = \sum\limits_{j=1}^{n} sign(\mathbf{g_j})^\intercal r \; \forall \mathbf{r} \in \mathbb{R}^d$. Applying the above results here, we get

\begin{equation}
    \sum\limits_{j=1}^{n} sign(\mathbf{g_j})^\intercal \mathbf{r_i} = \frac{\epsilon d}{\sqrt{k}} \; \forall 1 \leq i \leq k
\end{equation}

Now, similar to~\cite{tramer2017ensemble}, we compute the perturbation magnitude along a random permutation of the signed gradient. For each $1 \leq j \leq n$ and $1 \leq i \leq k$, we get :

\begin{equation}
    \begin{split}
        \mathbb{E}[\mathbf{g_j}^\intercal \mathbf{r_i}] &= \mathbb{E}\left[\sum\limits_{p=1}^{d} |g_j^{(p)}|\cdot sign(g_j^{(p)}) \cdot r_i^{(p)}\right] \\
        &= \sum\limits_{p=1}^{d} |g_j^{(p)}| \mathbb{E}\left[ sign(g_j^{(p)}) \cdot r_i^{(p)}\right] \\
        &= \frac{\epsilon ||\mathbf{g_j}||_1}{n\sqrt{k}}
    \end{split}
\end{equation}

\subsection{Additional Evaluation Details}\label{ap:eval}

\subsubsection{Details for Attacks}\label{ap:attacks}
We consider the following attacks for our evaluation. These attacks constitute the entire ensemble of attacks used in the AutoAttack Benchmark. We tune the attacks where necessary to get the strongest attack setting.

\textbf{APGD-CE/CW~\cite{croce2020reliable}} is a step-size free variant of the standard PGD attack. It adjusts the step size during the attack based on the convergence of the loss and the overall perturbation budget. We optimize the adaptive PGD-attack on the Cross Entropy (CE) and Carlini Wagner (CW) loss functions. We use the same set of parameters for the attack as mentioned in AutoAttack Benchmark other than the step size decay parameter $\alpha$ which we set to $0.1$ instead of $2$.

\textbf{FAB~\cite{croce2020minimally}} is a iterative first order attack that utilizes geometry of the decision boundary to minimize the perturbation required to cause mis-classification. We use the same set of parameters as AutoAttack.

\textbf{Square~\cite{andriushchenko2020square}} is an efficient black-box attack that uses random square shaped updates to approximate the decision boundary. We use the same set of parameters as AutoAttack.

\subsubsection{Evaluation on CIFAR10}\label{ap:cifar}
Table~\ref{tab:cifar_results} presents results for attacking the D3-S(4) ensemble and AT baseline, for the CIFAR10 classification task. While D3-S(4) offers some improvements at the smallest perturbation budgets, it quickly drops off. We suspect that a more carefully chosen feature space with redundancies for animal/vehicle classification would improve these results.

\begin{table}[H]
\small
\centering
\caption{CIFAR10 adversarial accuracies for $\ell_\infty$ and $\ell_2$ perturbation attacks for same configurations as in Table~\ref{tab:eval_l2}.}\label{tab:cifar_results}
\vspace{0.05in}
\begin{tabular}{@{}cccc@{}}
\toprule
\multicolumn{1}{l}{Attacks} & $\epsilon$ & \begin{tabular}[c]{@{}l@{}}\name (SIZE=1)\end{tabular} & \name (SIZE=4)  \\
\midrule
\multicolumn{4}{c}{$\ell_\infty$} \\ 
\midrule
\multicolumn{1}{l}{\multirow{4}{*}{\begin{tabular}[c]{@{}l@{}}APGD-\\CE(50)\end{tabular}}} & 1/255   & 1.7              & \textbf{59.8}    \\
\multicolumn{1}{l}{}                              & 4/255   & 0.1 & \textbf{1.2} \\
\multicolumn{1}{l}{}                              & 8/255   & \textbf{0.1} & 0 \\
\multicolumn{1}{l}{}                              & 16/255  & \textbf{0.1} & 0 \\\midrule
\multicolumn{4}{c}{$\ell_2$} \\                                               \midrule
\multicolumn{1}{l}{\multirow{4}{*}{\begin{tabular}[c]{@{}l@{}}APGD-\\CE(50)\end{tabular}}} & 0.5  & 15.1 & \textbf{63.5}    \\
\multicolumn{1}{l}{}                              & 1  & 13.6 & \textbf{39.7} \\
\multicolumn{1}{l}{}                              & 5  & 0.9 & \textbf{8.2} \\
\multicolumn{1}{l}{}                              & 10 & 0.8 & \textbf{6.2} \\ \bottomrule   
\end{tabular}
\end{table}

\end{document}